\documentclass[a4paper]{article}

\usepackage{INTERSPEECH2020}
\usepackage[table,xcdraw]{xcolor}
\usepackage{float}
\usepackage{comment}
\usepackage{multirow,tabularx}
\title{Transferring Monolingual Model to Low-Resource Language: The Case of Tigrinya}
\name{Abrhalei Tela$^1$, Abraham Woubie$^2$, Ville Hautam\"aki$^1$}
\address{
  $^1$School of Computing, University of Eastern Finland, Joensuu, Finland\\
  $^2$Department of Signal Processing and Acoustics, Aalto University, Espoo, Finland}
\email{abrht@uef.fi, abraham.zewoudie@aalto.fi, villeh@cs.uef.fi}

\begin{document}

\maketitle
\begin{abstract}

In recent years, transformer models have achieved great success in {\em natural language processing} (NLP) tasks. Most of the current state-of-the-art NLP results are achieved by using monolingual transformer models, where the model is pre-trained using a single language unlabelled text corpus. Then, the model is fine-tuned to the specific downstream task. However, the cost of pre-training a new transformer model is high for most languages. In this work, we propose a cost-effective transfer learning method to adopt a strong source language model, trained from a large monolingual corpus to a low-resource language. Thus, using XLNet language model, we demonstrate competitive performance with mBERT and a pre-trained target language model on the {\em cross-lingual sentiment} (CLS) dataset and on a new sentiment analysis dataset for low-resourced language Tigrinya. With only 10k examples of the given Tigrinya sentiment analysis dataset, English XLNet has achieved 78.88\% F1-Score outperforming BERT and mBERT by 10\% and 7\%, respectively. More interestingly, fine-tuning (English) XLNet model on the CLS dataset has promising results compared to mBERT and even outperformed mBERT for one dataset of the Japanese language.
  
\end{abstract}
\section{Introduction}

{\em Natural language processing} (NLP)~\cite{bb1055551717171} problems like machine translation~\cite{DBLPjournalscorrWuSCLNMKCGMKSJL16}, sentiment analysis~\cite{LiuSA}, and question answering~\cite{Rajpurkar_2016} have achieved great success with the emergence of transformer models~\cite{vaswani2017attention,devlin2018bert,yang2019xlnet}, and availability of large corpora and introduction of modern computing infrastructures. Compared to the traditional neural network methods, transformer models achieve not only lower error rates but also reduce the training time required on down streaming tasks, which makes them easier to be used by a wide range of applications.

However, most languages (especially the low-resource languages) in the world have limited available corpora~\cite{ruder2019unsupervised} to train language-specific transformer models~\cite{vaswani2017attention} from scratch. Training such a model from scratch can also be quite expensive in terms of computational power used~\cite{journalscorrabs190409408}. Thus, the similar explosion of state-of-the-art NLP models than in English language has not been materialized for many other languages. Then, naturally, we would like to find a way how to push these NLP models for multiple languages in a cost-effective manner. To tackle this problem, researchers have proposed multilingual transformer models such as mBERT~\cite{devlin2018bert} and XLM~\cite{lample2019crosslingual}. These models share a common vocabulary of multiple languages and pre-trained on a large text corpus of the given set of languages tokenized using the shared vocabulary. 
The multilingual transformer models have helped to push the state-of-the-art results on cross-lingual NLP tasks \cite{devlin2018bert,lample2019crosslingual,conneau2019unsupervised}. However, most multilingual models have performance trade-off between low and high-resource languages~\cite{conneau2019unsupervised}. High-resource languages dominate the performance of such models, but it usually under-performs when compared to the monolingual models~\cite{vries2019bertje, virtanen2019multilingual}. Moreover, only~$\sim$100 languages are used for pre-training such models, which can be ineffective for other unrepresented languages~\cite{virtanen2019multilingual}.

It was hypothesized in~\cite{k2019crosslingual} that the lexical overlap between different languages has a negligible role while the structural similarities, like morphology and word order, among languages to have a crucial role in cross-lingual success. In this work, our approach is to transfer a monolingual transformer model into a new target language. We transfer the source model at the lexical level by learning the target language's token embeddings. Our work provides additional evidence that strong monolingual representations are a useful initialization for cross-lingual transfer in line with~\cite{artetxe2019crosslingual}.

We show that monolingual models on language A can learn about language B without any shared vocabulary or shared pre-training data. This gives us new insights on using transformer models trained on single language to be fine-tuned using a labeled dataset of new unseen target languages. Furthermore, this helps the low-resource languages to use a monolingual transformer model pre-trained with high-resource language's text corpus. By using this approach, we can eliminate the cost of pre-training a new transformer model from scratch. Moreover, we empirically examined the ability of BERT, mBERT, and XLNet to generalize on a new target language. Based on our experiments, XLNet model can generalize more on new target languages. Finally, we publish the first publicly available sentiment analysis dataset for the Tigrinya language.







\section{Tigrinya}

Tigrinya is a language commonly used in Eritrea and Ethiopia, with more than 7 million speakers worldwide~\cite{Tedla2018MorphologicalSW}. It is one of the Semitic languages with the likes of Amharic, Arabic, and Hebrew~\cite{semiticlanguages}. While these languages have received a reasonably good focus from the NLP research community, Tigrinya has been one of the under-studied languages in NLP with no publicly available datasets and tools for NLP tasks such as machine translation, question answering, and sentiment analysis.

\begin{figure*}[h]
\begin{center}
  \includegraphics[width=0.8\textwidth,height=0.5\textheight,keepaspectratio]{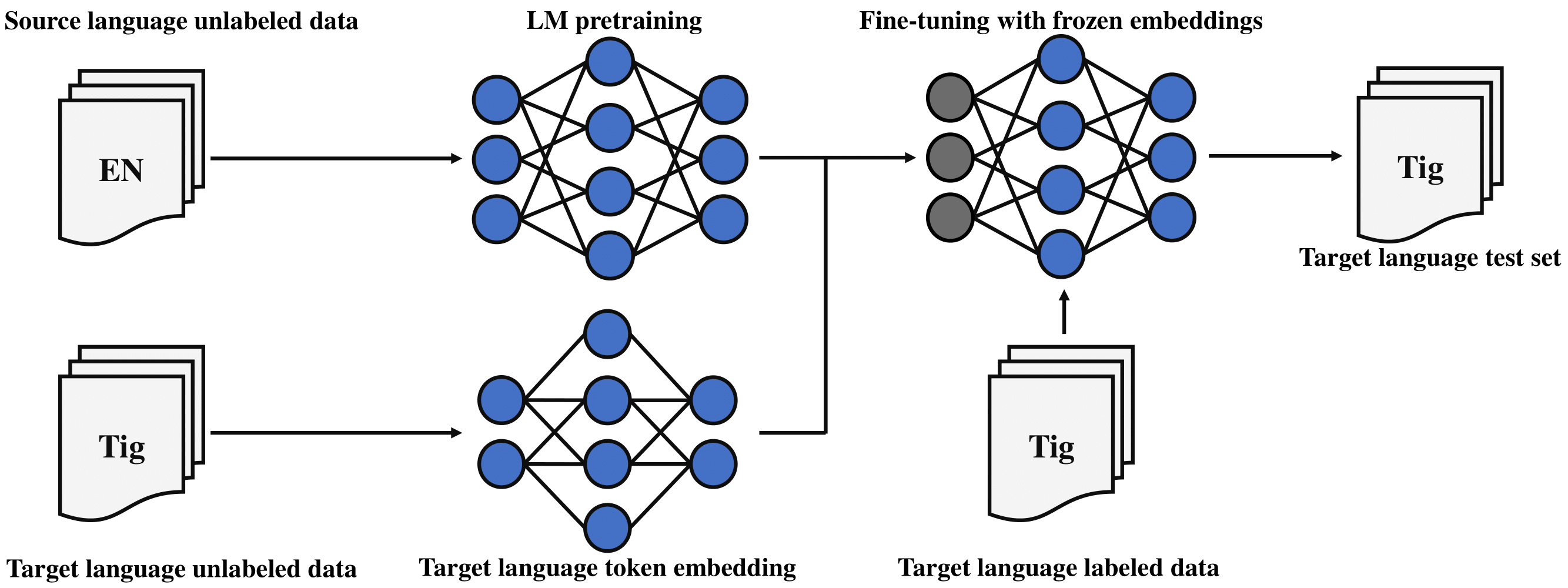}
  \caption{Transfer a monolingual transformer model to new target language }
  \vspace{-6.5ex}
  \label{fig21}
\end{center}
\end{figure*}

Tigrinya has its own alphabet chart called “Fidel” with some letters shared by other languages such as Amharic and Tigre~\cite{osman-mikami-2012-stemming}. The writing system is derived from Ge’ez language, in which each letter has a syllable of “consonant + vowel”, except in rare cases~\cite{semiticlanguages, mebrahtuEthio}. Tigrinya alphabet has 35 base letters with 7 vowels and some extra letters formed by variation of those base letters with 5 vowels to form a list of around 275 unique symbols~\cite{tigPOSyemane}. By default, Tigrinya has {\em subject-object-verb} (SOV) word order, although this is not always the case with some exception~\cite{Tedla2018MorphologicalSW, amanual}. The writing system is from left to right with each word is separated by a space; however, Tigrinya is a morphologically rich language where multiple morphemes can be packed together in a single word.


The morphology of Tigrinya is similar to Semitic languages, including Amharic and Arabic, which have a root-and-pattern type~\cite{Tedla2018MorphologicalSW}. Moreover, Tigrinya has a complex morphological structure with many variations of a given root verb not only by its prefix and suffix but also its internal inflections of the root-and-pattern type~\cite{Tedla2018MorphologicalSW, osman-mikami-2012-stemming}. For example, from a given root word “terefe” (fail), we can change its internal morphemes to have other new forms “terifu” (he failed) and “terifa” (she failed). Such gender differences make structural change in the root word. Besides this, Tigrinya has conjunctions and prepositions as part of the word itself.

Tedla and Yamamoto~\cite{Tedla2018MorphologicalSW} studied a detailed morphological segmentation of Tigrinya language. The authors chose to use Latin transliterations of the given Ge’ez scripts due to the syllabic properties of Tigrinya’s letters, which can result in alterations of characters in the segmentation boundaries. However, in this work, we have used the natural Ge’ez script text for our sentiment analysis task. A language-independent tokenizer, SentencePiece~\cite{kudo2018sentencepiece}, is trained using a large Tigrinya text corpus used for training our TigXLNet model to segment a natural Ge’ez script based Tigrinya text input.

\section{Cross-lingual transformer models}

\subsection{Background}

Multilingual transformer models are designed to have one common model representing multiple languages and then fine-tune on a downstream task of those languages~\cite{devlin2018bert,lample2019crosslingual,conneau2019unsupervised}. Multilingual BERT uses the same {\em masked language model} (MLM)~\cite{devlin2018bert} objective used for monolingual BERT trained using multiple languages. XLM, in contrast, tries to leverage parallel data by proposing a {\em translation language model}  (TLM)~\cite{lample2019crosslingual}. XLM-R~\cite{conneau2019unsupervised} has pushed the state-of-the-art results in many cross-lingual tasks by following the approach used by XLM, but scaling up the amount of training data and uncovering the low-resource vs. high-resource trade-off.

On the other hand, Chi et al.~\cite{chi2019monolingual} proposed a teacher-student framework based fine-tuning technique on a new target language’s text classification task, while Artetxe et al.~\cite{artetxe2019crosslingual} proposed a zero-shot shot based fine-tuning method to transfer a monolingual model into a new target language. Those zero-shot techniques are relatively cost-effective with less or zero numbers of labeled data required on target languages. However, none of them uses {\em permutation language model} (PLM)~\cite{yang2019xlnet} based XLNet, which, based on our experiment, could lead to better performance for unseen languages.

\subsection{Language model objectives}

MLM is an auto-encoding based pre-training language modeling objective, in which the model is trained to predict a set of corrupted tokens represented by “[MASK]” from a given sentence. From a given set of tokens of an input sentence with size $T$,
$x$ = [$x_1$, $x_2$, $x_3$, ... $x_T$], BERT first masks some tokens, $y$ = [$y_1$, $y_2$, $y_3$, ... $y_N$], of the total given tokens where N $<$ T. Then the learning objective will be to predict the masked tokens back with:  \[ \max_{\theta} \quad  \log p_\theta(y|x) =  \sum_{t=1}^{N}\, {\log p_\theta(y_t|x)}  \]  
TLM objective is an extension for MLM to take advantage of the parallel corpus for multilingual language representations. While the mathematical formulation is kept the same with MLM, TLM has more contextual information to learn from during pre-training. PLM is auto-regressive language modeling, which has access to bi-directional context while keeping the nature of auto-regressive language modeling. This way, PLM tries to resolve the limitations of MLM, independence assumption, and input noise pointed out by Yang et al.~\cite{yang2019xlnet}.

\subsection{Proposed method}

In this work, we have proposed a new transfer learning approach to use an already existing English monolingual transformer model to tackle downstream tasks of other unseen target languages. Hence, the language model pre-training for the source language is not a necessary step, making the proposed method more cost-efficient. The transformer models considered as source model in this work are BERT, XLNet, and mBERT. Figure~\ref{fig21} shows the graphical illustration of the proposed method.

To transfer a monolingual transformer model into a new target language, we followed three different steps. First, we generate a vocabulary for the target language using SentencePiece model trained on the language's unlabelled dataset. Then, we train a context-independent Word2Vec~\cite{mikolov2013efficient} based token embeddings for the vocabulary generated in the previous step. Finally, the given transformer model is fine-tuned on a labeled dataset of the target language with frozen token embeddings. By freezing the token embeddings of the model during fine-tuning, the transformer model can preserve the learned embeddings. This is necessary because the embedding technique used is a context-independent token embedding; however, in practice, it does not seem to have more performance difference.



\section{Experimental Setup and Result}

In this work, we conducted our experiment for the Tigrinya sentiment classification task on a newly created Tigrinya sentiment analysis dataset. Furthermore, we have tested our experiment on one of the standard cross-lingual datasets for sentiment analysis, the {\em cross-lingual sentiment} (CLS) dataset~\cite{Prettenhofer2010}.

\subsection{Datasets}

We have constructed a sentiment analysis dataset
for Tigrinya with two classes as positive and negative. The data has been collected from YouTube comments of Eritrean and Ethiopian music videos and short movie channels. It consists of around 30k automatically labeled training set, and two professionals had labeled the test set independently and considered only when they gave the same label to form the final 4k test examples with 2k positive and 2k negative. Additionally, we have used the CLS dataset for testing our proposed method on languages like German, French, and Japanese. It consists of English, German, French and Japanese languages collected from Amazon reviews on three different domains (Music, Books, and DVD).  

Text augmentation methods such as~\cite{sugiyama-yoshinaga-2019-data, Wei2019EDAED} are shown to increase the performance of text classification tasks with less available data. Back translation based data augmentation proposed by Sugiyama and Yoshinaga~\cite{sugiyama-yoshinaga-2019-data} requires a good machine translation model which is not always available for low-resourced languages like Tigrinya. Alternatively, Wei and Zou~\cite{Wei2019EDAED} proposed a natural but effective data augmentation using four different operations: synonym replacement, random swap, random insertion, and random deletion. We follow a similar approach by Wei and Zou~\cite{Wei2019EDAED}; however, we have used Word2Vec embeddings-based synonym replacement. In all our experiments, we have used the human-labeled 4k dataset as our test set while the augmented~$\sim$50k dataset for training unless otherwise stated.

\subsection{Baseline Models}

When evaluating our proposed method using the CLS dataset, we used mBERT fine-tuned on the same sized training data. This way, we can examine the ability of XLNet to understand new languages during fine-tuning comparatively with the mBERT trained on 104 languages, including the four languages of the CLS dataset. For Tigrinya sentiment analysis, we have evaluated the proposed method against a new transformer model, TigXLNet, which is purely pre-trained in a single Tigrinya language text corpus, then used to fine-tune on our new dataset. Furthermore, we have compared the generalization of BERT, XLNet, and mBERT on unseen target language, Tigrinya, with different configurations.

\subsection{Results on Tigrinya Sentiment Analysis Dataset}

As shown in Table~\ref{table41}, fine-tuning XLNet on Tigrinya sentiment analysis dataset is comparable to the result of fine-tuning TigXLNet and better than mBERT fine-tuned on the same dataset. Using the English XLNet model as the source transformer model, the proposed method has achieved 81.62\% in F1-Score, while the same method using mBERT source model has a lower F1-Score at 77.51\%.  Furthermore, the effectiveness of initializing the model with language-dependent embeddings instead of using random embeddings (source language embeddings) is also presented in the Table~\ref{table41}.  Both XLNet and mBERT under-performed when using random token embeddings compared to their corresponding models initialized with Word2Vec token embeddings. This shows that transferring a monolingual

\begin{center}
\begin{table}[ht!]
\captionsetup{justification=centering}
\caption{\label{tab:accents}Fine-tuning TigXLNet, mBERT and XLNet using Tigrinya sentiment analysis dataset.}
\centering
\begin{tabular}{ |c|c|c|c| } 

\cline{1-3}
\textbf{Models} & \textbf{Embedding} & \textbf{F1-Score} \\
\cline{1-3}
TigXLNet & - & \textbf{83.29} \\
\cline{1-3}
\multirow{2}{4em}{mBERT} & +random token embed. & 76.01  \\ 
\cline{2-3}
& +word2vec token embed. & 77.51 \\ 
\hline
\multirow{2}{4em}{XLNet} & +random token embed. & 77.83  \\ 
\cline{2-3}
& +word2vec token embed. & 81.62 \\ 
\hline
\end{tabular}
\label{table41}
\vspace{-0.7cm}
\end{table}
\end{center}

\noindent XLNet model into a new language like Tigrinya can result in a good performance at a little cost without needing to train language-specific transformer models.

\subsection{XLNet Frozen Weights}

We have tested the performance of XLNet model on Tigrinya sentiment analysis with different configurations, as presented in Table~\ref{table42}. In the first setup, we randomize the pre-trained XLNet model weights to examine if the performance we gain in an unseen language is from the learned XLNet weights, not just from the XLNet neural network architecture, and its ability to learn new features during fine-tuning. As we may expect, the model's performance decreases drastically compared to the model started with the pre-trained XLNet weights. The XLNet model initialized with randomized weights results in 53.93\% F1-Score, which is close to a result of a random model trained on binary classification tasks. In the second configuration, when all the transformer layers of XLNet are frozen during fine-tuning, the performance of the model has increased significantly from the previous configuration of randomly initialized weights by~$\sim$15\% (F1-Score). From these results, we can conclude that XLNet (English) model (initialized with random weights) cannot learn from the given labeled dataset at the fine-tuning stage. On the other hand, the pre-trained weights of XLNet have a general understanding of unseen languages like Tigrinya. Lastly, by fine-tuning XLNet on a labeled dataset of the target language, the model's performance gets better.

\begin{center}
\begin{table}[ht!]
\caption{\label{tab:accents} Fine-tuning XLNet using Tigrinya sentiment analysis dataset with different settings.}
\centering
\small
\begin{tabular}{ |c|c|c|c| } 

\cline{1-3}
\textbf{Model} & \textbf{Settings} & \textbf{F1-Score} \\
\cline{1-3}
\multirow{3}{4em}{XLNet} & +Random XLNet weights & 53.93  \\
\cline{2-3}
& +Frozen XLNet weights  & 68.14 \\ 
\cline{2-3}
& +Fine-tune XLNet weights &\textbf{81.62} \\ 
\hline
\end{tabular}
\label{table42}
\vspace{-0.7cm}
\end{table}
\end{center}

\subsection{Result on CLS Dataset}

In this experiment, monolingual XLNet is compared with mBERT. As in Table~\ref{table43}, monolingual XLNet pre-trained using English text corpus has abstract representations of other unseen languages such as German, French, and Japanese. Although the F1-Score of mBERT is expected to be higher for all datasets of those languages (mBERT pre-training language set includes those CLS languages), XLNet has achieved comparable results,

\begin{center}
\begin{table*}[ht]
\caption{
F1-Score on CLS dataset, note that we have used the same hyper-parameters and same dataset size for all models (all train and the unprocessed dataset of CLS is used for training, and the model is evaluated on the given test set)
}
\centering
\resizebox{\textwidth}{!}{%
\small{}
\begin{tabularx}{\textwidth}{|*{14}{c|}}
\cline{1-14}
\multirow{2}{*}{\textbf{Models}} 
  &\multicolumn{3}{c|}{\textbf{English}}
  &\multicolumn{3}{c|}{\textbf{German}} 
  &\multicolumn{3}{c|}{\textbf{French}}
  &\multicolumn{3}{c|}{\textbf{Japanese}}  
  & \multirow{2}{*}{\textbf{Avg.}} \\
\cline{2-13}
    &Books &DVD &Music
    &Books &DVD &Music  
    &Books &DVD &Music 
    &Books &DVD &Music & \\
\cline{1-14}
XLNet   & \scriptsize{\textbf{92.90}} &\scriptsize{\textbf{93.31}}  & \scriptsize{\textbf{92.02}} 
        & \scriptsize{85.23}         & \scriptsize{83.30}    & \scriptsize{83.89}
        & \scriptsize{73.05}         & \scriptsize{69.80}    & \scriptsize{70.12}
        & \scriptsize{83.20}         & \scriptsize{\textbf{86.07}}    & \scriptsize{85.24} & \scriptsize{83.08}  \\
\cline{1-14}
mBERT   & \scriptsize{92.78}           & \scriptsize{90.30}    & \scriptsize{91.88}
        & \scriptsize{\textbf{88.65}}           & \scriptsize{\textbf{85.85}}    & \scriptsize{\textbf{90.38}}
        & \scriptsize{\textbf{91.09}}           & \scriptsize{\textbf{88.57}}    & \scriptsize{\textbf{93.67}}
        & \scriptsize{\textbf{84.35}}           & \scriptsize{81.77}    & \scriptsize{\textbf{87.53}}  & \scriptsize{\textbf{88.90}} \\
\cline{1-14}
\end{tabularx}}
\label{table43}
\end{table*}
\end{center}
\vspace{-0.84cm}

\vspace{-2pt}
\noindent especially with German and Japanese dataset. Furthermore, XLNet even outperforms mBERT in one of the experiments for the Japanese language. From the results of this, we can deduce that XLNet is strong enough, compared to mBERT, to learn about unseen languages during fine-tuning of the new target language.

\subsection{BERT vs. XLNet on New Language }
In this experiment, as presented in Table~\ref{table44}, MLM based BERT and mBERT are compared to PLM based XLNet on a new target language: Tigrinya. By freezing all parameters of BERT, mBERT, and XLNet, except corresponding embedding and final linear layers, we can observe that BERT and mBERT are close to a random model with binary classification task. While frozen XLNet model results in more than 10\% F1-Score increase to both BERT and mBERT. This clearly shows that the pre-trained weights of XLNet have better generalization ability on unseen target language compared to both BERT and mBERT pre-trained weights. Furthermore, the positive effect of initializing all models with language-specific token embeddings can be observed from the Table~\ref{table44}. By initializing BERT and mBERT pre-trained models with Word2Vec token embeddings, the performance on fine-tuning Tigrinya sentiment analysis dataset has increased by around 2\% (F1-Score) when compared to their corresponding pre-trained models with random weights. Finally, we can observe that PLM based XLNet has outperformed MLM based BERT and mBERT in all different settings.

\begin{center}
\begin{table}[h]
\caption{Comparison of BERT, mBERT, and XLNet models fine-tuned using the Tigrinya sentiment analysis dataset. All hyper-parameters are the same for all models, including a learning rate of 2e-5, batch size of 32, the sequence length of 180, and 3 number of epochs.}
\centering
\resizebox{20em}{!}{%
\small
\begin{tabular}{ |c|c|c|c| } 

\cline{1-3}
\textbf{Models} & \textbf{Configuration} & \textbf{F1-Score} \\
\cline{1-3}
\multirow{3}{4em}{BERT} & +Frozen BERT weights & 54.91  \\
\cline{2-3}
& +Random embeddings &	74.26 \\ 
\cline{2-3}
& +Frozen token embeddings & 76.35 \\ 
\hline
\cline{1-3}
\multirow{3}{4em}{mBERT} & +Frozen mBERT weights	& 57.32  \\
\cline{2-3}
& +Random embeddings  & 76.01 \\ 
\cline{2-3}
& +Frozen token embeddings & 77.51 \\ 
\hline
\cline{1-3}
\multirow{3}{4em}{XLNet} & +Frozen XLNet weights &\textbf{68.14}  \\
\cline{2-3}
& +Random embeddings  &\textbf{77.83} \\ 
\cline{2-3}
& +Frozen token embeddings &\textbf{81.62} \\ 
\hline
\end{tabular}
}
\label{table44}
\vspace{-0.7cm}
\end{table}
\end{center}

\subsection{Effect of Dataset Size }
\begin{figure}[h]
\centering
\includegraphics[width=\columnwidth]{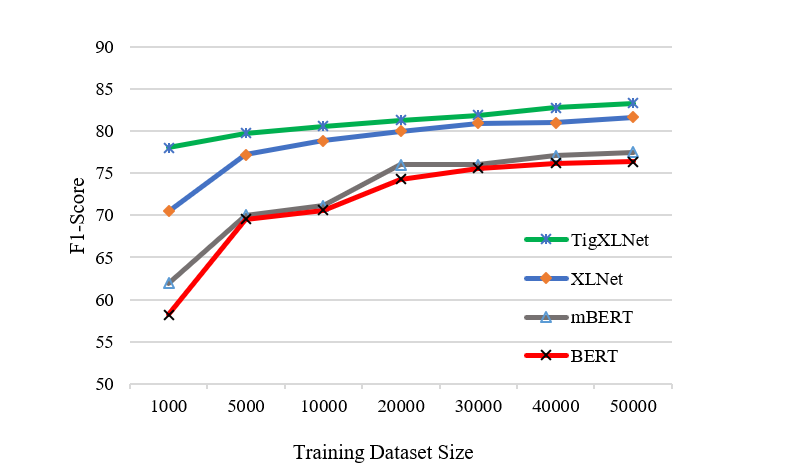}
\caption{The effect of dataset size for fine-tuning XLNet, TigXLNet, BERT, and mBERT on the Tigrinya sentiment analysis dataset.}
\label{fig42}
\vspace{-3ex}
\end{figure}

Figure~\ref{fig42} shows the effects of training dataset size for the performance of BERT, mBERT, XLNet, and TigXLNet based on the Tigrinya sentiment analysis dataset. By randomly selecting 1k, 5k, 10k, 20k, 30k, 40k, and full dataset of $\sim$50k examples, the performance of XLNet is dominant when compared to BERT and mBERT. All hyper-parameters of the models during fine-tuning are stayed fixed for all models except for TigXLNet, where the number of epochs is one as it tends to overfit if the number of epochs is larger. XLNet has achieved an F1-Score of 77.19\% with just 5k training examples while BERT and mBERT require full dataset size ($\sim$50k examples) to achieve 76.35\% and 77.51\% respectively. The performance of both XLNet and TigXLNet has increased by less than 3\%, with a dataset increase of 40k (10k to 50k). Based on this experiment, around 10k training examples could be enough to get a comparably good XLNet model fine-tuned for new language (Tigrinya) text classification tasks. Finally, with~$\sim$2 hours of fine-tuning XLNet using Google Colab's GPU, we can save the computational cost of pre-training TigXLNet from scratch, which takes 7 days using TPU v3-8 of 8 cores and 128GB memory.

\section{Conclusion}
In this research, we have performed an empirical study on the ability of XLNet to generalize on a new language. Interestingly, transferring an English XLNet model to a new target language, Tigrinya, we achieve comparable performance to a monolingual XLNet model (TigXLNet) pre-trained on Tigrinya text corpus. Computational saving of performing transfer learning only is enormous. 
The proposed method also has comparable results to mBERT on CLS dataset, especially on Japanese and German languages. In our experiment, it is shown that PLM based XLNet has better performance in the case of unseen languages when compared to MLM based BERT and mBERT. Furthermore, we have released a new Tigrinya sentiment analysis dataset and a new XLNet model specifically for Tigrinya language, TigXLNet, which could help NLP downstream tasks of Tigrinya.
Our experiment results also hint that training multilingual transformer models using PLM could achieve better performance boost across a range of downstream NLP tasks. This is due to the advantages of PLM over other language models like MLM to discover more insights about languages that are not even in the pre-training corpus. 

\bibliographystyle{IEEEtran}

\bibliography{paper}


\end{document}